\documentclass[11pt,a4paper]{article}

\usepackage[utf8]{inputenc}
\usepackage[T2A,T1]{fontenc}
\usepackage[ukrainian,english]{babel}
\usepackage{amsmath,amssymb}
\usepackage{graphicx}
\usepackage{booktabs,multirow,array,tabularx}
\usepackage{hyperref}
\usepackage{natbib}
\usepackage[margin=1in]{geometry}
\usepackage{microtype}
\usepackage{enumitem}
\usepackage{caption,subcaption}
\usepackage{float}
\usepackage{xcolor}
\usepackage{tikz}
\usepackage{pgfplots}
\pgfplotsset{compat=1.18}
\usepgfplotslibrary{colorbrewer}

\newcolumntype{R}[1]{>{\raggedleft\arraybackslash}p{#1}}

\hypersetup{
    colorlinks=true,
    linkcolor=blue!70!black,
    citecolor=blue!70!black,
    urlcolor=blue!70!black,
}

\title{UA-Legal-Bench: A Benchmark for Evaluating Large Language Models on Ukrainian Legal Reasoning}

\author{
  Volodymyr Ovcharov \\
  SecondLayer \\
  \texttt{vladimir@legal.org.ua} \\
  \url{https://legal.org.ua}
}

\date{}

\begin{document}
\maketitle

\begin{abstract}
Legal NLP benchmarks are overwhelmingly English-centric, leaving failure modes in morphologically rich, non-Latin-script languages undetected. We introduce \textbf{UA-Legal-Bench}, a five-task benchmark for evaluating large language models on Ukrainian legal reasoning, built from the Unified State Register of Court Decisions (EDRSR) -- one of the world's largest open judicial corpora (99.5 million decisions). The benchmark comprises: \textbf{(1)}~case-type classification (4 classes, $n{=}2{,}000$), \textbf{(2)}~judgment form classification (4 classes, $n{=}2{,}000$), \textbf{(3)}~case-outcome prediction (6 classes, $n{=}800$), \textbf{(4)}~legal norm extraction ($n{=}1{,}794$), and \textbf{(5)}~cause category prediction (22 classes, $n{=}1{,}871$). We evaluate 11 LLMs (3B--675B) from five families under zero-shot and 3-shot prompting via AWS Bedrock with 158K API calls. Our results reveal sharply task-dependent few-shot effects: few-shot prompting improves judgment form classification by up to $+38.6$~pp but has mixed effects on outcome prediction. We show that accuracy is misleading on imbalanced legal tasks: the model with highest COP accuracy (62\%) is a majority-class predictor (macro-F1: 23\%), while the genuinely best model scores only 44\% macro-F1. Within-family scaling analysis reveals that 8B models can match frontier performance on surface-level tasks but scaling thresholds vary dramatically across families. We release all data, prompts, and model predictions.

\noindent\textbf{Data:} \url{https://huggingface.co/datasets/overthelex/ua-legal-bench}
\end{abstract}


\section{Introduction}
\label{sec:intro}

The rapid adoption of large language models (LLMs) in legal practice has spurred the development of benchmarks to evaluate their reasoning capabilities. LegalBench~\citep{guha2024legalbench}, LexGLUE~\citep{chalkidis2022lexglue}, and CUAD~\citep{hendrycks2021cuad} have established rigorous evaluation standards -- yet all are English-only. Multilingual efforts such as LEXTREME~\citep{niklaus2023lextreme} and MultiLegalPile~\citep{niklaus2024multilegal} extend coverage to EU languages but exclude Cyrillic-script jurisdictions and civil-law systems outside Western Europe.

This gap is consequential. Ukrainian, with its Cyrillic script, agglutinative morphology, and seven grammatical cases, presents a fundamentally different tokenization surface. Our prior work~\citep{ovcharov2025fertility} shows that subword tokenizers fragment Ukrainian legal text at rates 1.6$\times$ higher than English equivalents across frontier models, directly inflating inference costs and degrading in-context learning. Moreover, the Ukrainian legal system -- a civil-law jurisdiction with codified statutes rather than common-law precedent -- requires distinct reasoning patterns that English benchmarks do not capture.

Ukraine's Unified State Register of Court Decisions (EDRSR) provides an unparalleled empirical foundation: 99.5 million decisions published since 2006, constituting one of the largest open judicial datasets in the world. We leverage this resource to construct UA-Legal-Bench, a benchmark spanning five tasks that probe progressively deeper levels of legal understanding:

\begin{enumerate}[nosep]
    \item \textbf{Case-Type Classification (CTC)} -- classifying decisions into four jurisdictional categories;
    \item \textbf{Judgment Form Classification (JFC)} -- identifying the document type (decision, resolution, sentence, ruling);
    \item \textbf{Case-Outcome Prediction (COP)} -- predicting the judicial ruling from case facts alone;
    \item \textbf{Norm Extraction (NE)} -- extracting legal norm citations from decision text;
    \item \textbf{Cause Category Prediction (CCP)} -- classifying the legal subject matter into 22 macro-categories.
\end{enumerate}

\noindent Our contributions are:

\begin{enumerate}[nosep]
    \item The first comprehensive legal reasoning benchmark for a Cyrillic-script, civil-law jurisdiction, with 2,000 decisions spanning five tasks of increasing difficulty.
    \item A standardized evaluation of 11 LLMs (3B--675B, five families) with 158K API calls under both zero-shot and few-shot conditions.
    \item The discovery of sharply task-dependent few-shot effects, including few-shot as a scale equalizer ($+38.6$~pp for a 12B model, approaching 120B performance).
    \item Within-family scaling analysis showing that the parameter threshold for Ukrainian legal reasoning varies dramatically across model families.
    \item All data, prompts, model predictions, and evaluation code released publicly.
\end{enumerate}

\section{Related Work}
\label{sec:related}

\paragraph{English legal benchmarks.}
LegalBench~\citep{guha2024legalbench} defines 162 tasks spanning six categories of legal reasoning, evaluated on frontier LLMs. LexGLUE~\citep{chalkidis2022lexglue} provides a multi-task benchmark for legal language understanding across seven datasets. CUAD~\citep{hendrycks2021cuad} focuses on contract review. More recently, \citet{katz2024gpt} demonstrate that GPT-4 passes the Uniform Bar Examination, highlighting the rapid capability growth that benchmarks must track. These benchmarks have driven substantial progress but evaluate exclusively in English and predominantly within common-law systems.

\paragraph{Multilingual legal NLP.}
LEXTREME~\citep{niklaus2023lextreme} aggregates 11 datasets across 24 languages in the EU legal domain, while SCALE~\citep{niklaus2024scale} adds complexity-scaled tasks. MultiLegalPile~\citep{niklaus2024multilegal} provides a 689~GB multilingual legal corpus spanning 24 EU languages. \citet{chalkidis2020legal} provide foundational pre-trained models for legal text, and \citet{zheng2021does} examine cross-lingual transfer in legal NLP. Despite broad coverage, these efforts exclude Ukrainian and do not address the distinctive challenges of Cyrillic-script legal text.

\paragraph{Few-shot evaluation and metric choice.}
\citet{brown2020language} establish few-shot prompting as a standard evaluation paradigm, but its effectiveness on non-English, domain-specific tasks remains understudied. \citet{zhao2021calibrate} show that few-shot performance is sensitive to example selection and ordering, motivating our use of fixed, stratified examples. The problem of metric selection for imbalanced classification is well-studied~\citep{grandini2020metrics}; we show that this concern is especially acute in legal benchmarks where majority-class prediction can yield high accuracy but low macro-F1.

\paragraph{Ukrainian NLP.}
No prior benchmark exists for Ukrainian legal reasoning. \citet{romanyshyn2023unlp} organize the first shared task on Ukrainian NLP at EACL 2023, covering NER and lemmatization but not legal tasks. General Ukrainian NLP resources remain limited compared to EU official languages: Ukrainian constitutes only 0.5\% of the mC4 corpus, 18$\times$ less than Russian and 2.4$\times$ less than Polish~\citep{ovcharov2025crosslingual}. This data scarcity compounds with tokenizer inefficiency~\citep{rust2021good} to create a ``double penalty'' for Ukrainian legal NLP.

\paragraph{Positioning.}
UA-Legal-Bench fills this gap as the first legal benchmark for a Cyrillic-script, civil-law jurisdiction (Table~\ref{tab:comparison}). Unlike prior benchmarks that typically evaluate a single task type, UA-Legal-Bench spans five tasks of increasing difficulty and explicitly evaluates the interaction between few-shot prompting, model scale, and metric choice.

\begin{table}[t]
\centering
\caption{Comparison with existing legal NLP benchmarks.}
\label{tab:comparison}
\small
\begin{tabular}{l c c c c c}
\toprule
& \textbf{Legal-} & \textbf{Lex-} & \textbf{LEX-} & \textbf{SCALE} & \textbf{UA-Legal-} \\
& \textbf{Bench} & \textbf{GLUE} & \textbf{TREME} & & \textbf{Bench} \\
\midrule
Languages    & 1 (en) & 1 (en) & 24 (EU) & 5 (EU) & 1 (uk) \\
Cyrillic     & \texttimes & \texttimes & \texttimes & \texttimes & \checkmark \\
Tasks        & 162 & 7 & 11 & 5 & 5 \\
Models eval. & 20 & 7 & 6 & 8 & 11 \\
Scale range  & frontier & BERT & BERT & BERT & 3B--675B \\
Few-shot     & \checkmark & \texttimes & \texttimes & \texttimes & \checkmark \\
Scaling anal.& \texttimes & \texttimes & \texttimes & \texttimes & \checkmark \\
Corpus size  & varied & 35K & 50K & 50K & 99.5M \\
\bottomrule
\end{tabular}
\end{table}

\section{Benchmark Design}
\label{sec:design}

\subsection{Data Source}
\label{sec:data}

All tasks draw from the Unified State Register of Court Decisions (EDRSR, \foreignlanguage{ukrainian}{Єдиний державний реєстр судових рішень}), an official Ukrainian government registry containing 99.5 million court decisions spanning 2006--2026 across all jurisdictional levels and court types. Each record includes structured metadata (court, date, case number, jurisdiction type, judgment form, cause category) and the full decision text.

We sample 2,000 decisions from the 2024 partition: 500 per jurisdiction (civil, criminal, commercial, administrative), stratified across judgment forms. Only substantive decisions with full text between 2,000 and 30,000 characters are included (median length: 12,060 characters). For each decision, we extract: the facts section (for COP), cited legal norms (for NE), and the ruling outcome (for COP), using rule-based parsers validated against EDRSR's canonical document structure.

\subsection{Tasks}
\label{sec:tasks}

\paragraph{Task 1: Case-Type Classification (CTC).}
Given a court decision, classify it into one of four jurisdictional categories: \emph{civil} (\foreignlanguage{ukrainian}{цивільне}), \emph{criminal} (\foreignlanguage{ukrainian}{кримінальне}), \emph{commercial} (\foreignlanguage{ukrainian}{господарське}), or \emph{administrative} (\foreignlanguage{ukrainian}{адміністративне}). Labels derive from EDRSR metadata. This task evaluates basic legal document understanding.

\emph{Metric:} Accuracy. \emph{Size:} $n{=}2{,}000$.

\paragraph{Task 2: Judgment Form Classification (JFC).}
Classify the document type: \emph{rishennya} (decision), \emph{postanova} (resolution), \emph{vyrok} (sentence), or \emph{ukhvala} (ruling). Unlike CTC, this requires understanding the procedural nature of the document -- a resolution and a decision may come from the same court but serve different legal functions.

\emph{Metric:} Accuracy. \emph{Size:} $n{=}2{,}000$.

\paragraph{Task 3: Case-Outcome Prediction (COP).}
Given only the facts section of a court decision (ruling masked), predict the outcome from six classes: \emph{granted}, \emph{partial}, \emph{denied}, \emph{closed}, \emph{guilty}, \emph{left without consideration}. This task requires legal reasoning over factual circumstances and is restricted to decisions where the facts section and outcome label can be reliably extracted.

\emph{Metric:} Macro-F1 (due to 61\% majority class). \emph{Size:} $n{=}800$.

\paragraph{Task 4: Norm Extraction (NE).}
Given a court decision, extract all legal norm references (e.g., ``\foreignlanguage{ukrainian}{ст.~625 ЦК України}'', ``\foreignlanguage{ukrainian}{ч.~2 ст.~16 ЦПК України}''). Ground truth is constructed by parsing canonical citation patterns from decision texts, yielding a mean of 7.5 norms per document.

\emph{Metric:} Set-level F1. \emph{Size:} $n{=}1{,}794$.

\paragraph{Task 5: Cause Category Prediction (CCP).}
Classify the legal subject matter of the case into one of 22 macro-categories (e.g., \emph{contracts}, \emph{theft}, \emph{family}, \emph{pension/social}, \emph{property}, \emph{violence}), derived from EDRSR's 4,106-category taxonomy via keyword-based aggregation.

\emph{Metric:} Accuracy. \emph{Size:} $n{=}1{,}871$.

\subsection{Evaluation Protocol}
\label{sec:protocol}

All tasks are evaluated under zero-shot and 3-shot prompting. Prompts are in Ukrainian with standardized instruction templates. All inference uses temperature~0 for reproducibility. Few-shot examples are drawn from a fixed pool (not overlapping with test data) using a deterministic seed, with stratified sampling to cover label diversity. For imbalanced tasks (COP: 61\% majority class; CCP: 31\% majority class), we report macro-F1 rather than accuracy to prevent majority-class bias. 95\% Wilson confidence intervals are $\pm$0.8~pp for CTC/JFC ($n{=}2{,}000$), $\pm$3.4~pp for COP ($n{=}800$), and $\pm$2.3~pp for CCP ($n{=}1{,}871$).

\section{Models}
\label{sec:models}

We evaluate eleven LLMs accessed via AWS Bedrock from five model families, with three families (Mistral, Meta, NVIDIA) represented at multiple scales to enable within-family scaling analysis (Table~\ref{tab:models}).

\begin{table}[t]
\centering
\caption{Models evaluated. Frontier models (top) and their smaller counterparts (bottom) enable within-family scaling analysis. Fertility: tokens per word on Ukrainian legal text~\citep{ovcharov2025fertility}.}
\label{tab:models}
\begin{tabular}{llrr}
\toprule
\textbf{Family} & \textbf{Model} & \textbf{Params} & \textbf{Fertility} \\
\midrule
\multirow{3}{*}{Mistral}  & Mistral Large 3   & 675B/41B & 3.06 \\
                           & Ministral 8B      & 8B       & -- \\
                           & Ministral 3B      & 3B       & -- \\
\midrule
\multirow{3}{*}{Meta}     & Llama 4 Maverick  & 400B/17B & 2.43 \\
                           & Llama 3.3 70B     & 70B      & 2.65 \\
                           & Llama 3.1 8B      & 8B       & -- \\
\midrule
\multirow{2}{*}{NVIDIA}   & Nemotron Super 3  & 120B/12B & 3.08 \\
                           & Nemotron Nano 12B & 12B      & -- \\
\midrule
Amazon                     & Nova Pro          & undiscl. & 3.61 \\
\multirow{2}{*}{Qwen}     & Qwen3 235B        & 235B/22B & 3.89 \\
                           & Qwen3 32B         & 32B      & 3.90 \\
\bottomrule
\end{tabular}
\end{table}

The seven frontier models span 32B to 675B total parameters with tokenizer fertility ranging from 2.43 (Llama 4 Maverick) to 3.90 (Qwen3 32B) -- a 1.6$\times$ spread meaning identical Ukrainian legal text consumes 60\% more tokens on the least efficient tokenizer. The four smaller models (3B--12B) from the Mistral, Meta, and NVIDIA families enable direct measurement of how performance scales within a model family on Ukrainian legal tasks.

\section{Results}
\label{sec:results}

We report results from 158,419 API calls across 11 models. Table~\ref{tab:main} presents performance across all five tasks, with frontier and smaller models separated.

\begin{table}[t]
\centering
\caption{UA-Legal-Bench results. CTC/JFC/CCP: accuracy (\%). COP: macro-F1 (\%), which accounts for the imbalanced label distribution (61\% ``granted''). NE: set-level F1. Best zero-shot result per task in bold. Baselines: majority = always predict most frequent class; random = uniform random.}
\label{tab:main}
\small
\begin{tabular}{l R{0.7cm} R{0.7cm} R{0.7cm} R{0.7cm} R{0.7cm} R{0.7cm} R{0.7cm} R{0.7cm} R{0.7cm} R{0.7cm}}
\toprule
& \multicolumn{2}{c}{\textbf{CTC}} & \multicolumn{2}{c}{\textbf{JFC}} & \multicolumn{2}{c}{\textbf{COP}} & \multicolumn{2}{c}{\textbf{NE}} & \multicolumn{2}{c}{\textbf{CCP}} \\
& \multicolumn{2}{c}{\emph{acc.}} & \multicolumn{2}{c}{\emph{acc.}} & \multicolumn{2}{c}{\emph{m-F1}} & \multicolumn{2}{c}{\emph{F1}} & \multicolumn{2}{c}{\emph{acc.}} \\
\cmidrule(lr){2-3} \cmidrule(lr){4-5} \cmidrule(lr){6-7} \cmidrule(lr){8-9} \cmidrule(lr){10-11}
\textbf{Model} & ZS & FS & ZS & FS & ZS & FS & ZS & FS & ZS & FS \\
\midrule
\emph{Majority}     & 25.0 & -- & 47.3 & -- & 12.7 & -- & -- & -- & 30.9 & -- \\
\midrule
\multicolumn{11}{l}{\emph{Frontier models}} \\
Maverick    & 96.4 & 97.4 & 75.8 & 90.8 & 32.5 & 38.9 & .379 & .381 & 50.5 & 52.7 \\
Llama 3.3   & 97.3 & 97.7 & 74.1 & \textbf{91.8} & \textbf{40.6} & 36.6 & .372 & .375 & 43.9 & 49.1 \\
Mistral L3  & \textbf{97.4} & \textbf{97.9} & 74.5 & 89.3 & 35.4 & 39.2 & .368 & .366 & 47.9 & 51.7 \\
Nemotron S3 & \textbf{97.4} & 97.5 & \textbf{84.0} & 92.6 & 22.7 & 26.9 & .365 & .369 & 49.2 & 52.6 \\
Nova Pro    & 97.2 & 97.2 & 79.5 & 90.8 & 39.4 & \textbf{43.8} & \textbf{.383} & \textbf{.391} & 46.5 & 52.1 \\
Qwen3 235B  & \textbf{97.4} & 98.0 & 76.3 & 91.1 & 34.7 & 34.3 & .374 & .380 & \textbf{51.1} & \textbf{55.5} \\
Qwen3 32B   & 97.1 & 97.5 & 76.1 & 76.5 & 31.4 & 27.3 & .339 & .358 & 48.3 & 50.5 \\
\midrule
\multicolumn{11}{l}{\emph{Smaller models (within-family scaling)}} \\
Ministral 8B   & 97.5 & 97.3 & 69.0 & 82.5 & 31.4 & 34.7 & .350 & .345 & 40.6 & 44.4 \\
Ministral 3B   & 95.5 & 92.6 & 74.9 & 75.8 & 27.4 & 21.5 & .353 & .351 & 45.9 & 46.5 \\
Nemotron N.~12B & 84.2 & 96.9 & 51.8 & 90.4 & 19.3 & 23.6 & .318 & .299 & 40.4 & 41.8 \\
Llama 3.1 8B   & 86.6 & 88.8 & 36.5 & 65.7 & 5.3  & 14.0 & .359 & .347 & 23.6 & 23.4 \\
\bottomrule
\end{tabular}
\end{table}

\subsection{Task Difficulty Gradient}
\label{sec:gradient}

\begin{figure}[t]
\centering
\begin{tikzpicture}[x=1pt,y=1pt]
\definecolor{fillColor}{RGB}{255,255,255}
\path[use as bounding box,fill=fillColor,fill opacity=0.00] (0,0) rectangle (325.21,202.36);
\begin{scope}
\path[clip] (  0.00,  0.00) rectangle (325.21,202.36);
\definecolor{fillColor}{RGB}{255,255,255}

\path[fill=fillColor] (  0.00,  0.00) rectangle (325.21,202.36);
\end{scope}
\begin{scope}
\path[clip] ( 29.55, 14.91) rectangle (320.72,165.40);
\definecolor{drawColor}{gray}{0.92}

\path[draw=drawColor,line width= 0.5pt,line join=round] ( 29.55, 21.75) --
	(320.72, 21.75);

\path[draw=drawColor,line width= 0.5pt,line join=round] ( 29.55, 56.63) --
	(320.72, 56.63);

\path[draw=drawColor,line width= 0.5pt,line join=round] ( 29.55, 91.52) --
	(320.72, 91.52);

\path[draw=drawColor,line width= 0.5pt,line join=round] ( 29.55,126.40) --
	(320.72,126.40);

\path[draw=drawColor,line width= 0.5pt,line join=round] ( 29.55,161.29) --
	(320.72,161.29);
\definecolor{fillColor}{RGB}{233,196,106}

\path[fill=fillColor] (268.92, 21.75) rectangle (285.72, 94.37);
\definecolor{fillColor}{RGB}{69,123,157}

\path[fill=fillColor] (288.52, 21.75) rectangle (305.32, 89.01);
\definecolor{fillColor}{RGB}{233,196,106}

\path[fill=fillColor] (156.93, 21.75) rectangle (173.73, 70.80);
\definecolor{fillColor}{RGB}{69,123,157}

\path[fill=fillColor] (176.53, 21.75) rectangle (193.33, 70.28);
\definecolor{fillColor}{RGB}{233,196,106}

\path[fill=fillColor] ( 44.94, 21.75) rectangle ( 61.74,157.96);
\definecolor{fillColor}{RGB}{69,123,157}

\path[fill=fillColor] ( 64.54, 21.75) rectangle ( 81.34,157.33);
\definecolor{fillColor}{RGB}{233,196,106}

\path[fill=fillColor] (100.94, 21.75) rectangle (117.74,145.94);
\definecolor{fillColor}{RGB}{69,123,157}

\path[fill=fillColor] (120.54, 21.75) rectangle (137.33,129.45);
\definecolor{fillColor}{RGB}{233,196,106}

\path[fill=fillColor] (212.93, 21.75) rectangle (229.72, 73.96);
\definecolor{fillColor}{RGB}{69,123,157}

\path[fill=fillColor] (232.52, 21.75) rectangle (249.32, 73.18);
\definecolor{drawColor}{RGB}{0,0,0}

\path[draw=drawColor,line width= 0.5pt,line join=round] (274.52, 99.14) --
	(280.12, 99.14);

\path[draw=drawColor,line width= 0.5pt,line join=round] (277.32, 99.14) --
	(277.32, 90.32);

\path[draw=drawColor,line width= 0.5pt,line join=round] (274.52, 90.32) --
	(280.12, 90.32);

\path[draw=drawColor,line width= 0.5pt,line join=round] (294.12, 93.12) --
	(299.72, 93.12);

\path[draw=drawColor,line width= 0.5pt,line join=round] (296.92, 93.12) --
	(296.92, 83.05);

\path[draw=drawColor,line width= 0.5pt,line join=round] (294.12, 83.05) --
	(299.72, 83.05);

\path[draw=drawColor,line width= 0.5pt,line join=round] (162.53, 82.88) --
	(168.13, 82.88);

\path[draw=drawColor,line width= 0.5pt,line join=round] (165.33, 82.88) --
	(165.33, 59.38);

\path[draw=drawColor,line width= 0.5pt,line join=round] (162.53, 59.38) --
	(168.13, 59.38);

\path[draw=drawColor,line width= 0.5pt,line join=round] (182.13, 78.44) --
	(187.73, 78.44);

\path[draw=drawColor,line width= 0.5pt,line join=round] (184.93, 78.44) --
	(184.93, 53.40);

\path[draw=drawColor,line width= 0.5pt,line join=round] (182.13, 53.40) --
	(187.73, 53.40);

\path[draw=drawColor,line width= 0.5pt,line join=round] ( 50.54,158.56) --
	( 56.14,158.56);

\path[draw=drawColor,line width= 0.5pt,line join=round] ( 53.34,158.56) --
	( 53.34,157.37);

\path[draw=drawColor,line width= 0.5pt,line join=round] ( 50.54,157.37) --
	( 56.14,157.37);

\path[draw=drawColor,line width= 0.5pt,line join=round] ( 70.14,157.66) --
	( 75.74,157.66);

\path[draw=drawColor,line width= 0.5pt,line join=round] ( 72.94,157.66) --
	( 72.94,156.26);

\path[draw=drawColor,line width= 0.5pt,line join=round] ( 70.14,156.26) --
	( 75.74,156.26);

\path[draw=drawColor,line width= 0.5pt,line join=round] (106.54,151.01) --
	(112.14,151.01);

\path[draw=drawColor,line width= 0.5pt,line join=round] (109.34,151.01) --
	(109.34,128.50);

\path[draw=drawColor,line width= 0.5pt,line join=round] (106.54,128.50) --
	(112.14,128.50);

\path[draw=drawColor,line width= 0.5pt,line join=round] (126.14,139.03) --
	(131.73,139.03);

\path[draw=drawColor,line width= 0.5pt,line join=round] (128.94,139.03) --
	(128.94,125.15);

\path[draw=drawColor,line width= 0.5pt,line join=round] (126.14,125.15) --
	(131.73,125.15);

\path[draw=drawColor,line width= 0.5pt,line join=round] (218.53, 76.31) --
	(224.13, 76.31);

\path[draw=drawColor,line width= 0.5pt,line join=round] (221.33, 76.31) --
	(221.33, 71.73);

\path[draw=drawColor,line width= 0.5pt,line join=round] (218.53, 71.73) --
	(224.13, 71.73);

\path[draw=drawColor,line width= 0.5pt,line join=round] (238.12, 75.21) --
	(243.72, 75.21);

\path[draw=drawColor,line width= 0.5pt,line join=round] (240.92, 75.21) --
	(240.92, 69.00);

\path[draw=drawColor,line width= 0.5pt,line join=round] (238.12, 69.00) --
	(243.72, 69.00);
\end{scope}
\begin{scope}
\path[clip] (  0.00,  0.00) rectangle (325.21,202.36);
\definecolor{drawColor}{gray}{0.30}

\node[text=drawColor,anchor=base east,inner sep=0pt, outer sep=0pt, scale=  0.72] at ( 25.50, 19.27) {0};

\node[text=drawColor,anchor=base east,inner sep=0pt, outer sep=0pt, scale=  0.72] at ( 25.50, 54.15) {25};

\node[text=drawColor,anchor=base east,inner sep=0pt, outer sep=0pt, scale=  0.72] at ( 25.50, 89.04) {50};

\node[text=drawColor,anchor=base east,inner sep=0pt, outer sep=0pt, scale=  0.72] at ( 25.50,123.92) {75};

\node[text=drawColor,anchor=base east,inner sep=0pt, outer sep=0pt, scale=  0.72] at ( 25.50,158.81) {100};
\end{scope}
\begin{scope}
\path[clip] (  0.00,  0.00) rectangle (325.21,202.36);
\definecolor{drawColor}{gray}{0.30}

\node[text=drawColor,anchor=base,inner sep=0pt, outer sep=0pt, scale=  0.72] at ( 63.14,  5.90) {CTC};

\node[text=drawColor,anchor=base,inner sep=0pt, outer sep=0pt, scale=  0.72] at (119.14,  5.90) {JFC};

\node[text=drawColor,anchor=base,inner sep=0pt, outer sep=0pt, scale=  0.72] at (175.13,  5.90) {COP};

\node[text=drawColor,anchor=base,inner sep=0pt, outer sep=0pt, scale=  0.72] at (231.12,  5.90) {NE};

\node[text=drawColor,anchor=base,inner sep=0pt, outer sep=0pt, scale=  0.72] at (287.12,  5.90) {CCP};
\end{scope}
\begin{scope}
\path[clip] (  0.00,  0.00) rectangle (325.21,202.36);
\definecolor{drawColor}{RGB}{0,0,0}

\node[text=drawColor,rotate= 90.00,anchor=base,inner sep=0pt, outer sep=0pt, scale=  0.90] at ( 10.70, 90.16) {Score (\% or F1$\times$100)};
\end{scope}
\begin{scope}
\path[clip] (  0.00,  0.00) rectangle (325.21,202.36);
\definecolor{fillColor}{RGB}{233,196,106}

\path[fill=fillColor] (125.62,179.48) rectangle (138.91,192.77);
\end{scope}
\begin{scope}
\path[clip] (  0.00,  0.00) rectangle (325.21,202.36);
\definecolor{fillColor}{RGB}{69,123,157}

\path[fill=fillColor] (177.20,179.48) rectangle (190.49,192.77);
\end{scope}
\begin{scope}
\path[clip] (  0.00,  0.00) rectangle (325.21,202.36);
\definecolor{drawColor}{RGB}{0,0,0}

\node[text=drawColor,anchor=base west,inner sep=0pt, outer sep=0pt, scale=  0.72] at (143.99,183.65) {Few-shot};
\end{scope}
\begin{scope}
\path[clip] (  0.00,  0.00) rectangle (325.21,202.36);
\definecolor{drawColor}{RGB}{0,0,0}

\node[text=drawColor,anchor=base west,inner sep=0pt, outer sep=0pt, scale=  0.72] at (195.57,183.65) {Zero-shot};
\end{scope}
\end{tikzpicture}
\caption{Task difficulty gradient across frontier models. Bars show mean score; whiskers show min--max range. JFC shows the largest few-shot gain; COP shows the widest model spread.}
\label{fig:difficulty}
\end{figure}
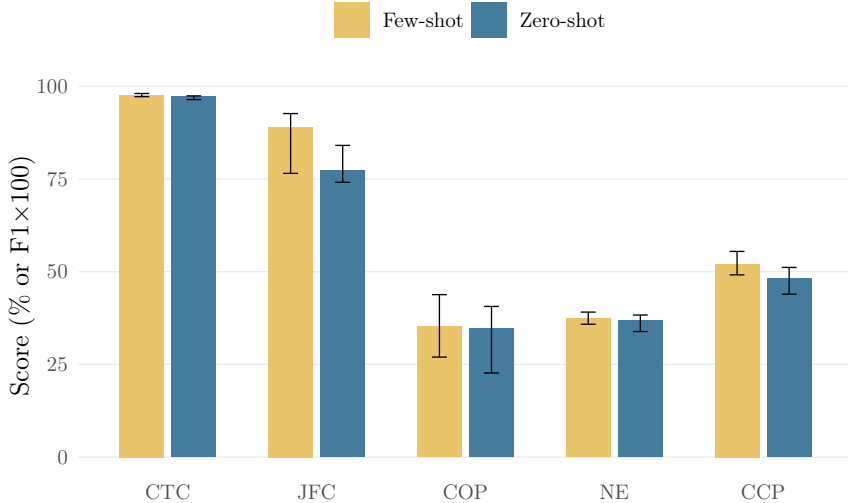

The five tasks form a clear difficulty gradient (Figure~\ref{fig:difficulty}). CTC is nearly solved: all frontier models exceed 96\% zero-shot, with only 1.0~pp separating best from worst. JFC is substantially harder (74--84\% ZS), requiring models to distinguish procedurally similar document types. COP is the hardest task: frontier models achieve only 23--41\% macro-F1 zero-shot, demanding genuine legal reasoning over masked outcomes. CCP sits between JFC and COP in difficulty (44--51\% accuracy ZS), testing domain knowledge of legal subject matter.

\subsection{Few-Shot Effects Are Task-Dependent}
\label{sec:fewshot}

\begin{figure}[t]
\centering
\input{figures/fig_fewshot_heatmap.tex}
\caption{Few-shot delta (pp) across all 11 models and 5 tasks. Green = few-shot helps, red = hurts. JFC shows consistent large gains; COP is mixed; NE is near-zero.}
\label{fig:heatmap}
\end{figure}

The central finding of UA-Legal-Bench is that few-shot prompting effects are sharply task-dependent and cannot be predicted from task difficulty alone (Figure~\ref{fig:heatmap}). Table~\ref{tab:delta} summarizes the frontier model deltas.

\begin{table}[t]
\centering
\caption{Few-shot delta (pp) relative to zero-shot for frontier models. COP uses macro-F1 delta. Bold: $|\Delta| > 5$~pp.}
\label{tab:delta}
\small
\begin{tabular}{l R{1.0cm} R{1.0cm} R{1.0cm} R{1.0cm} R{1.0cm}}
\toprule
\textbf{Model} & \textbf{CTC} & \textbf{JFC} & \textbf{COP} & \textbf{NE} & \textbf{CCP} \\
\midrule
Nova Pro          & 0.0  & \textbf{+11.3} & +4.4          & +.008 & \textbf{+5.6} \\
Maverick          & +1.0 & \textbf{+15.0} & \textbf{+6.4} & +.002 & +2.2 \\
Mistral L3        & +0.5 & \textbf{+14.8} & +3.8          & $-$.002 & +3.8 \\
Llama 3.3         & +0.4 & \textbf{+17.7} & $-$4.0        & +.003 & \textbf{+5.2} \\
Nemotron S3       & +0.1 & \textbf{+8.6}  & +4.2          & +.004 & +3.4 \\
Qwen3 235B        & +0.6 & \textbf{+14.8} & $-$0.4        & +.006 & +4.4 \\
Qwen3 32B         & +0.4 & +0.4           & $-$4.1        & +.019 & +2.2 \\
\bottomrule
\end{tabular}
\end{table}

\paragraph{JFC: Few-shot consistently helps.}
Six of seven models gain 8.6--17.7~pp from few-shot examples on judgment form classification. The largest gain is Llama 3.3 70B ($+17.7$~pp, from 74.1\% to 91.8\%). The sole exception is Qwen3 32B ($+0.4$~pp), which also shows the weakest few-shot response on COP. This suggests that JFC benefits from format learning: the four judgment forms have distinctive textual signatures that few-shot examples help models recognize.

\paragraph{COP: Few-shot effects vary; accuracy is misleading.}
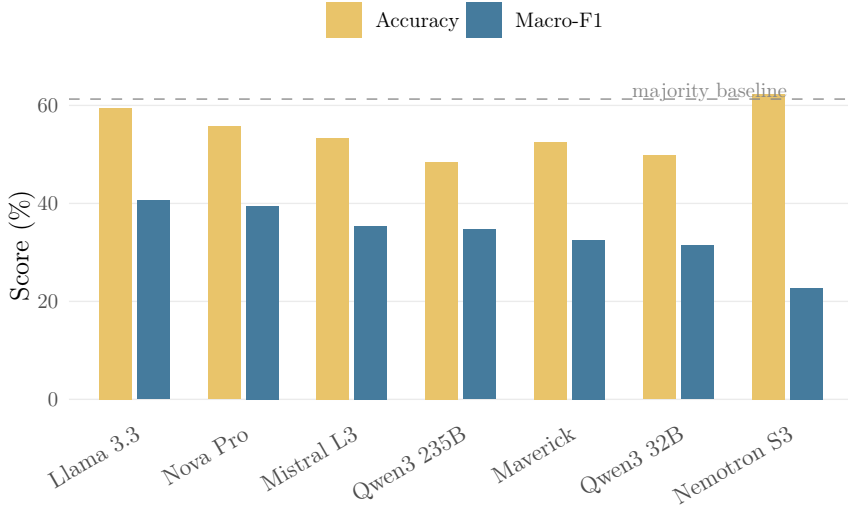
\begin{figure}[t]
\centering
\begin{tikzpicture}[x=1pt,y=1pt]
\definecolor{fillColor}{RGB}{255,255,255}
\path[use as bounding box,fill=fillColor,fill opacity=0.00] (0,0) rectangle (325.21,202.36);
\begin{scope}
\path[clip] (  0.00,  0.00) rectangle (325.21,202.36);
\definecolor{fillColor}{RGB}{255,255,255}

\path[fill=fillColor] (  0.00,  0.00) rectangle (325.21,202.36);
\end{scope}
\begin{scope}
\path[clip] ( 25.95, 37.67) rectangle (320.71,165.40);
\definecolor{drawColor}{gray}{0.92}

\path[draw=drawColor,line width= 0.5pt,line join=round] ( 25.95, 43.48) --
	(320.71, 43.48);

\path[draw=drawColor,line width= 0.5pt,line join=round] ( 25.95, 80.34) --
	(320.71, 80.34);

\path[draw=drawColor,line width= 0.5pt,line join=round] ( 25.95,117.20) --
	(320.71,117.20);

\path[draw=drawColor,line width= 0.5pt,line join=round] ( 25.95,154.07) --
	(320.71,154.07);
\definecolor{fillColor}{RGB}{233,196,106}

\path[fill=fillColor] ( 78.14, 43.48) rectangle ( 90.43,146.14);
\definecolor{fillColor}{RGB}{69,123,157}

\path[fill=fillColor] ( 92.47, 43.48) rectangle (104.76,116.10);
\definecolor{fillColor}{RGB}{233,196,106}

\path[fill=fillColor] ( 37.20, 43.48) rectangle ( 49.49,153.15);
\definecolor{fillColor}{RGB}{69,123,157}

\path[fill=fillColor] ( 51.53, 43.48) rectangle ( 63.82,118.31);
\definecolor{fillColor}{RGB}{233,196,106}

\path[fill=fillColor] (119.08, 43.48) rectangle (131.37,141.90);
\definecolor{fillColor}{RGB}{69,123,157}

\path[fill=fillColor] (133.41, 43.48) rectangle (145.70,108.73);
\definecolor{fillColor}{RGB}{233,196,106}

\path[fill=fillColor] (200.97, 43.48) rectangle (213.25,140.43);
\definecolor{fillColor}{RGB}{69,123,157}

\path[fill=fillColor] (215.29, 43.48) rectangle (227.58,103.38);
\definecolor{fillColor}{RGB}{233,196,106}

\path[fill=fillColor] (160.03, 43.48) rectangle (172.31,132.69);
\definecolor{fillColor}{RGB}{69,123,157}

\path[fill=fillColor] (174.35, 43.48) rectangle (186.64,107.44);
\definecolor{fillColor}{RGB}{233,196,106}

\path[fill=fillColor] (241.91, 43.48) rectangle (254.19,135.27);
\definecolor{fillColor}{RGB}{69,123,157}

\path[fill=fillColor] (256.23, 43.48) rectangle (268.52,101.35);
\definecolor{fillColor}{RGB}{233,196,106}

\path[fill=fillColor] (282.85, 43.48) rectangle (295.13,158.31);
\definecolor{fillColor}{RGB}{69,123,157}

\path[fill=fillColor] (297.17, 43.48) rectangle (309.46, 85.32);
\definecolor{drawColor}{RGB}{136,136,136}

\path[draw=drawColor,line width= 0.5pt,dash pattern=on 4pt off 4pt ,line join=round] ( 25.95,156.46) -- (320.71,156.46);

\node[text=drawColor,anchor=base east,inner sep=0pt, outer sep=0pt, scale=  0.71] at (296.15,157.15) {majority baseline};
\end{scope}
\begin{scope}
\path[clip] (  0.00,  0.00) rectangle (325.21,202.36);
\definecolor{drawColor}{gray}{0.30}

\node[text=drawColor,anchor=base east,inner sep=0pt, outer sep=0pt, scale=  0.72] at ( 21.90, 41.00) {0};

\node[text=drawColor,anchor=base east,inner sep=0pt, outer sep=0pt, scale=  0.72] at ( 21.90, 77.86) {20};

\node[text=drawColor,anchor=base east,inner sep=0pt, outer sep=0pt, scale=  0.72] at ( 21.90,114.73) {40};

\node[text=drawColor,anchor=base east,inner sep=0pt, outer sep=0pt, scale=  0.72] at ( 21.90,151.59) {60};
\end{scope}
\begin{scope}
\path[clip] (  0.00,  0.00) rectangle (325.21,202.36);
\definecolor{drawColor}{gray}{0.30}

\node[text=drawColor,rotate= 30.00,anchor=base east,inner sep=0pt, outer sep=0pt, scale=  0.80] at ( 53.27, 28.85) {Llama 3.3};

\node[text=drawColor,rotate= 30.00,anchor=base east,inner sep=0pt, outer sep=0pt, scale=  0.80] at ( 94.21, 28.85) {Nova Pro};

\node[text=drawColor,rotate= 30.00,anchor=base east,inner sep=0pt, outer sep=0pt, scale=  0.80] at (135.15, 28.85) {Mistral L3};

\node[text=drawColor,rotate= 30.00,anchor=base east,inner sep=0pt, outer sep=0pt, scale=  0.80] at (176.09, 28.85) {Qwen3 235B};

\node[text=drawColor,rotate= 30.00,anchor=base east,inner sep=0pt, outer sep=0pt, scale=  0.80] at (217.03, 28.85) {Maverick};

\node[text=drawColor,rotate= 30.00,anchor=base east,inner sep=0pt, outer sep=0pt, scale=  0.80] at (257.97, 28.85) {Qwen3 32B};

\node[text=drawColor,rotate= 30.00,anchor=base east,inner sep=0pt, outer sep=0pt, scale=  0.80] at (298.91, 28.85) {Nemotron S3};
\end{scope}
\begin{scope}
\path[clip] (  0.00,  0.00) rectangle (325.21,202.36);
\definecolor{drawColor}{RGB}{0,0,0}

\node[text=drawColor,rotate= 90.00,anchor=base,inner sep=0pt, outer sep=0pt, scale=  0.90] at ( 10.70,101.54) {Score (\%)};
\end{scope}
\begin{scope}
\path[clip] (  0.00,  0.00) rectangle (325.21,202.36);
\definecolor{fillColor}{RGB}{233,196,106}

\path[fill=fillColor] (122.95,179.48) rectangle (136.24,192.77);
\end{scope}
\begin{scope}
\path[clip] (  0.00,  0.00) rectangle (325.21,202.36);
\definecolor{fillColor}{RGB}{69,123,157}

\path[fill=fillColor] (175.41,179.48) rectangle (188.70,192.77);
\end{scope}
\begin{scope}
\path[clip] (  0.00,  0.00) rectangle (325.21,202.36);
\definecolor{drawColor}{RGB}{0,0,0}

\node[text=drawColor,anchor=base west,inner sep=0pt, outer sep=0pt, scale=  0.72] at (141.32,183.65) {Accuracy};
\end{scope}
\begin{scope}
\path[clip] (  0.00,  0.00) rectangle (325.21,202.36);
\definecolor{drawColor}{RGB}{0,0,0}

\node[text=drawColor,anchor=base west,inner sep=0pt, outer sep=0pt, scale=  0.72] at (193.78,183.65) {Macro-F1};
\end{scope}
\end{tikzpicture}
\caption{COP accuracy vs macro-F1 for frontier models. Nemotron achieves the highest accuracy (near the majority baseline) but the lowest macro-F1 -- it predicts ``granted'' for 97\% of cases. Nova Pro is the genuinely best model by macro-F1.}
\label{fig:cop_gap}
\end{figure}

COP has a heavily imbalanced label distribution (61\% ``granted''), making accuracy a poor metric (Figure~\ref{fig:cop_gap}). Nemotron Super 3 achieves the highest \emph{accuracy} (62.3\% ZS) but the lowest \emph{macro-F1} among frontier models (22.7\%). Per-class analysis reveals why: Nemotron recalls 97\% of ``granted'' cases but 0\% of ``guilty'' -- it is a majority-class predictor. In contrast, Nova Pro (macro-F1 39.4\% ZS, 43.8\% FS) achieves 86\% recall on ``guilty'', 62\% on ``granted'', and 43\% on ``denied'', demonstrating genuine cross-class reasoning. Few-shot prompting generally helps COP on macro-F1 (Maverick: 32.5\%$\to$38.9\%), with the notable exception of Llama 3.3 which \emph{drops} from 40.6\% to 36.6\%. This task demonstrates why imbalanced legal benchmarks require class-aware metrics.

\paragraph{CCP: Moderate consistent gains.}
Cause category prediction shows modest, consistently positive few-shot effects ($+2.2$ to $+5.6$~pp), suggesting that topic classification benefits from exemplar-based calibration without the negative length effects seen in COP.

\paragraph{CTC: Ceiling effect.}
Few-shot effects on CTC are minimal ($-0.0$ to $+1.0$~pp), reflecting a ceiling: when zero-shot accuracy already exceeds 96\%, examples add little.

\subsection{Model Rankings Vary by Task}
\label{sec:rankings}

No single model dominates across all tasks. Nemotron Super 3 leads on JFC (84.0\% ZS) but is a weak majority-class predictor on COP (mF1 22.7\%). Nova Pro leads on COP (mF1 39.4\% ZS) and NE (.383) but is mid-pack on JFC (79.5\%). Qwen3 235B leads on CCP (51.1\%) and ties for first on CTC (97.4\%). This task-dependent ranking instability means that evaluating on a single legal task -- as most prior benchmarks do -- provides an incomplete picture of model capability.

Larger models do not reliably outperform smaller ones: Qwen3 235B scores lower than Qwen3 32B on COP macro-F1 (34.7\% vs 31.4\%) and Ministral 8B (31.4\%) nearly matches its 675B sibling Mistral Large (35.4\%).

\subsection{Norm Extraction}
\label{sec:ne}

Norm extraction is evaluated via normalized (article, law-code) pair matching, yielding F1 scores of 0.318--0.391. Frontier models achieve F1 0.339--0.391 while smaller models are competitive (Llama 3.1 8B: 0.359, Ministral 3B: 0.353). This is the only task where small models match frontier performance, suggesting that norm extraction relies on pattern recognition rather than deep legal reasoning. Few-shot prompting has negligible effect on NE ($|\Delta\text{F1}| < 0.02$), contrasting sharply with its large effect on JFC.

\subsection{Scaling Analysis}
\label{sec:scaling}

\begin{figure}[t]
\centering
\input{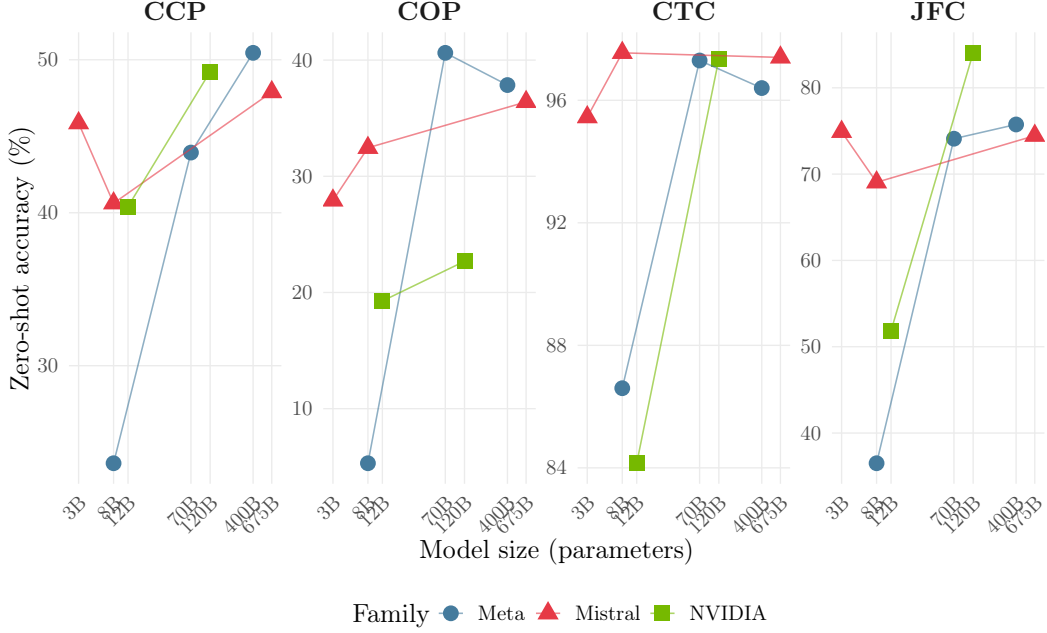}
\caption{Within-family scaling on four tasks (zero-shot). Mistral family (red) shows near-flat scaling on CTC and COP; Meta family (blue) shows steep scaling; NVIDIA (green) is intermediate.}
\label{fig:scaling}
\end{figure}

Table~\ref{tab:main} includes four smaller models (3B--12B) from three families, enabling within-family scaling analysis (Figure~\ref{fig:scaling}). The results reveal that scaling effects are highly task-dependent:

\paragraph{Mistral family (3B $\to$ 8B $\to$ 675B).}
On CTC, Ministral 8B (97.5\%) \emph{matches} Mistral Large 675B (97.4\%) -- an 84$\times$ parameter reduction with no accuracy loss. On COP macro-F1, Ministral 8B (31.4\%) approaches Mistral Large (35.4\%), a surprisingly small gap given the 84$\times$ scale difference. Only on JFC and CCP do larger models show clear advantages. Ministral 3B remains competitive on CTC (95.5\%) and JFC (74.9\%), but collapses on COP with few-shot prompting (mF1 21.5\%, $-5.9$~pp from ZS).

\paragraph{NVIDIA family (12B $\to$ 120B).}
Nemotron Nano 12B shows a scaling gap on COP macro-F1: 19.3\% vs Super's 22.7\%. Both models are weak on COP (near majority baseline), but the gap widens dramatically on CTC accuracy (84.2\% vs 97.4\%). However, few-shot prompting narrows the gap on JFC: Nano jumps from 51.8\% to 90.4\% ($+38.6$~pp), nearly matching Super's 92.6\%. This is the largest few-shot gain in our benchmark, suggesting that few-shot examples can partially compensate for a 10$\times$ parameter deficit on format-learnable tasks.

\paragraph{Meta family (8B $\to$ 70B $\to$ 400B).}
Llama 3.1 8B is the weakest model overall: COP mF1 5.3\% ZS, JFC 36.5\% ZS, CCP 23.6\% ZS. The jump to Llama 3.3 70B is dramatic (COP mF1: $+35.3$~pp, JFC: $+37.6$~pp), indicating that the Meta family requires substantially more parameters for Ukrainian legal reasoning than the Mistral family.

\paragraph{Implications.}
These results challenge the assumption that legal AI requires frontier-scale models. For simpler tasks (CTC, NE), 8B models suffice. For harder tasks (COP, CCP), scaling matters -- but the scaling curve varies dramatically by family: Mistral's 8B model is competitive while Meta's 8B model is not.

\section{Discussion}
\label{sec:discussion}

\paragraph{Task-dependent few-shot effects.}
Our central finding -- that few-shot prompting can simultaneously help and hurt the same model on different tasks -- has practical implications. Legal AI practitioners cannot assume that few-shot prompting will improve performance; the effect must be measured per task. We hypothesize that the direction of the effect depends on the ratio of \emph{format signal} (learnable from examples) to \emph{length cost} (prompt inflation from tokenizer fertility). JFC has high format signal (distinctive document headers), while COP has low format signal (outcomes depend on factual reasoning, not surface patterns).

\paragraph{Few-shot as scale equalizer.}
The Nemotron Nano result (JFC: 51.8\% $\to$ 90.4\% with few-shot) demonstrates that few-shot prompting can partially substitute for model scale on format-learnable tasks. This has cost implications: achieving 90\% JFC accuracy with a 12B model and few-shot examples is far cheaper than deploying a 120B model zero-shot.

\paragraph{Comparison with English benchmarks.}
On LegalBench's comparable tasks, frontier LLMs routinely exceed 90\% accuracy~\citep{guha2024legalbench}. Our CTC results (96--98\%) are consistent, but COP (23--41\% macro-F1) and CCP (44--55\%) reveal substantially more variance, suggesting that the difficulty gap is task-dependent rather than uniform across all legal reasoning.

\paragraph{Why Ukrainian is hard.}
Three factors compound: \textbf{(1)}~tokenizer inefficiency inflates prompt length, reducing effective context; \textbf{(2)}~morphological richness (7 cases, 3 genders, synthetic verb forms) creates surface variation invisible to English-trained models; \textbf{(3)}~civil-law reasoning patterns -- statute application rather than case precedent -- differ from common-law training data that dominates LLM pretraining corpora.

\begin{figure}[t]
\centering
\begin{tikzpicture}[x=1pt,y=1pt]
\definecolor{fillColor}{RGB}{255,255,255}
\path[use as bounding box,fill=fillColor,fill opacity=0.00] (0,0) rectangle (252.94,180.67);
\begin{scope}
\path[clip] (  0.00,  0.00) rectangle (252.94,180.67);
\definecolor{fillColor}{RGB}{255,255,255}

\path[fill=fillColor] (  0.00,  0.00) rectangle (252.94,180.67);
\end{scope}
\begin{scope}
\path[clip] ( 25.95, 15.61) rectangle (248.45,143.72);
\definecolor{drawColor}{gray}{0.92}

\path[draw=drawColor,line width= 0.5pt,line join=round] ( 25.95, 21.44) --
	(248.45, 21.44);

\path[draw=drawColor,line width= 0.5pt,line join=round] ( 25.95, 49.04) --
	(248.45, 49.04);

\path[draw=drawColor,line width= 0.5pt,line join=round] ( 25.95, 76.63) --
	(248.45, 76.63);

\path[draw=drawColor,line width= 0.5pt,line join=round] ( 25.95,104.23) --
	(248.45,104.23);

\path[draw=drawColor,line width= 0.5pt,line join=round] ( 25.95,131.83) --
	(248.45,131.83);
\definecolor{fillColor}{RGB}{255,153,0}

\path[fill=fillColor] ( 69.40, 21.44) rectangle ( 90.26,135.97);

\path[fill=fillColor] (138.93, 21.44) rectangle (159.79,121.34);

\path[fill=fillColor] (208.46, 21.44) rectangle (229.32,137.90);
\definecolor{fillColor}{RGB}{118,185,0}

\path[fill=fillColor] ( 45.07, 21.44) rectangle ( 65.93, 83.81);

\path[fill=fillColor] (114.60, 21.44) rectangle (135.46, 99.81);

\path[fill=fillColor] (184.13, 21.44) rectangle (204.99, 89.60);
\end{scope}
\begin{scope}
\path[clip] (  0.00,  0.00) rectangle (252.94,180.67);
\definecolor{drawColor}{gray}{0.30}

\node[text=drawColor,anchor=base east,inner sep=0pt, outer sep=0pt, scale=  0.72] at ( 21.90, 18.96) {0};

\node[text=drawColor,anchor=base east,inner sep=0pt, outer sep=0pt, scale=  0.72] at ( 21.90, 46.56) {10};

\node[text=drawColor,anchor=base east,inner sep=0pt, outer sep=0pt, scale=  0.72] at ( 21.90, 74.15) {20};

\node[text=drawColor,anchor=base east,inner sep=0pt, outer sep=0pt, scale=  0.72] at ( 21.90,101.75) {30};

\node[text=drawColor,anchor=base east,inner sep=0pt, outer sep=0pt, scale=  0.72] at ( 21.90,129.35) {40};
\end{scope}
\begin{scope}
\path[clip] (  0.00,  0.00) rectangle (252.94,180.67);
\definecolor{drawColor}{gray}{0.30}

\node[text=drawColor,anchor=base,inner sep=0pt, outer sep=0pt, scale=  0.80] at ( 67.66,  6.06) {Original (UK)};

\node[text=drawColor,anchor=base,inner sep=0pt, outer sep=0pt, scale=  0.80] at (137.20,  6.06) {English};

\node[text=drawColor,anchor=base,inner sep=0pt, outer sep=0pt, scale=  0.80] at (206.73,  6.06) {Detailed (UK)};
\end{scope}
\begin{scope}
\path[clip] (  0.00,  0.00) rectangle (252.94,180.67);
\definecolor{drawColor}{RGB}{0,0,0}

\node[text=drawColor,rotate= 90.00,anchor=base,inner sep=0pt, outer sep=0pt, scale=  0.90] at ( 10.70, 79.67) {COP Macro-F1 (\%)};
\end{scope}
\begin{scope}
\path[clip] (  0.00,  0.00) rectangle (252.94,180.67);
\definecolor{fillColor}{RGB}{118,185,0}

\path[fill=fillColor] ( 81.11,157.80) rectangle ( 94.40,171.09);
\end{scope}
\begin{scope}
\path[clip] (  0.00,  0.00) rectangle (252.94,180.67);
\definecolor{fillColor}{RGB}{255,153,0}

\path[fill=fillColor] (145.98,157.80) rectangle (159.27,171.09);
\end{scope}
\begin{scope}
\path[clip] (  0.00,  0.00) rectangle (252.94,180.67);
\definecolor{drawColor}{RGB}{0,0,0}

\node[text=drawColor,anchor=base west,inner sep=0pt, outer sep=0pt, scale=  0.72] at ( 99.48,161.97) {Nemotron S3};
\end{scope}
\begin{scope}
\path[clip] (  0.00,  0.00) rectangle (252.94,180.67);
\definecolor{drawColor}{RGB}{0,0,0}

\node[text=drawColor,anchor=base west,inner sep=0pt, outer sep=0pt, scale=  0.72] at (164.35,161.97) {Nova Pro};
\end{scope}
\end{tikzpicture}
\caption{COP macro-F1 across three prompt variants. Ukrainian prompts yield stable results; English prompt effects vary by model.}
\label{fig:prompt}
\end{figure}
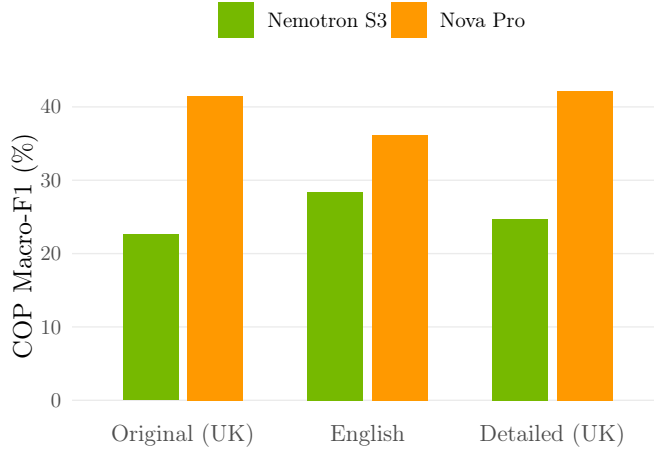

\paragraph{Prompt sensitivity.}
To test robustness, we evaluate COP with three prompt variants on two models (Figure~\ref{fig:prompt}): the original Ukrainian prompt, an English translation, and a detailed Ukrainian prompt with explicit role-playing. Results are stable across Ukrainian variants (Nova Pro: 41.5\% vs 42.2\% macro-F1, $\Delta{=}0.7$~pp; Nemotron: 22.6\% vs 24.7\%, $\Delta{=}2.1$~pp). English prompts cause moderate degradation for Nova Pro ($-5.3$~pp) but a surprising improvement for Nemotron ($+5.8$~pp), suggesting that prompt language interacts with model architecture in non-trivial ways.

\paragraph{Label quality.}
We validate ground truth labels by independently annotating 50 COP and 50 CCP decisions using Claude Sonnet 4.6 as an expert judge. COP label agreement is 70\% (regex-parsed outcome vs LLM judgment), and CCP agreement is 58\% (keyword-mapped category vs LLM classification). The moderate CCP agreement reflects the inherent ambiguity of topic classification -- many cases span multiple categories -- rather than systematic labeling errors.

\paragraph{Cross-year stability.}
To verify the benchmark is not specific to 2024 data, we sample 200 decisions from 2020 and evaluate CTC with two models. Results are stable: Nova Pro scores 94.5\% (2020) vs 97.2\% (2024, $\Delta{=}{-}2.7$~pp) and Nemotron scores 95.0\% vs 97.4\% ($\Delta{=}{-}2.4$~pp), confirming that the benchmark captures stable model capabilities rather than year-specific artifacts.

\paragraph{Statistical significance.}
McNemar's test confirms that the COP performance gap between Nova Pro and Nemotron Super 3 is significant ($\chi^2{=}9.2$, $p{=}0.002$), validating that the macro-F1 difference (39.4\% vs 22.7\%) reflects genuinely different prediction patterns rather than noise.

\paragraph{Limitations.}
The COP evaluation set ($n{=}800$) is smaller than other tasks due to the requirement for reliable facts extraction and outcome labeling, yielding wider confidence intervals ($\pm 3.4$~pp). COP label agreement of 70\% with an independent LLM judge suggests approximately 30\% noise in outcome labels, which may attenuate measured model differences. The benchmark currently covers a single jurisdiction; cross-lingual evaluation (e.g., Polish, Czech) is planned as future work.

\section{Conclusion}
\label{sec:conclusion}

We present UA-Legal-Bench, the first benchmark for evaluating LLMs on Ukrainian legal reasoning. Across five tasks and eleven models (158K evaluations, 3B--675B), we find that: (1)~case-type classification is nearly solved even at 8B scale, (2)~judgment form classification benefits dramatically from few-shot examples (up to $+38.6$~pp), (3)~case-outcome prediction reveals a critical metric choice: the highest-accuracy model is a majority-class predictor, while macro-F1 identifies genuinely capable models, (4)~norm extraction is the only task where small models match frontier performance, and (5)~the parameter threshold for competence varies dramatically by model family.

Two findings are particularly striking. First, accuracy is dangerously misleading on imbalanced legal tasks: Nemotron Super 3 achieves 62\% COP accuracy but macro-F1 of only 23\% (majority baseline: 13\%), while Nova Pro scores 44\% macro-F1 by actually distinguishing between outcome classes. Second, few-shot prompting can partially compensate for a 10$\times$ parameter deficit on format-learnable tasks (Nemotron Nano: $+38.6$~pp on JFC). These patterns -- invisible in English-only, accuracy-only evaluation -- underscore the need for multilingual legal benchmarks with appropriate metrics.

\paragraph{Data availability.}
All benchmark data, prompts, model predictions, and evaluation code are available at \url{https://huggingface.co/datasets/overthelex/ua-legal-bench} under CC-BY-4.0.

\section*{Ethics and Broader Impact}

UA-Legal-Bench is constructed entirely from publicly available court decisions published by the Ukrainian government under open access. All personal identifiers in EDRSR decisions are anonymized at source (e.g., OSOBA\_1, ADRESA\_1). The benchmark is intended for evaluating NLP models, not for automated judicial decision-making. We caution against using COP results to build predictive systems for real cases, as the task is designed as a benchmark probe, not a deployment-ready pipeline. The benchmark may contain biases inherent in the Ukrainian judicial system, including regional and temporal variation in case outcomes.

\section*{Datasheet for UA-Legal-Bench}

Following \citet{gebru2021datasheets}, we provide key dataset documentation.

\paragraph{Motivation.} UA-Legal-Bench was created to address the absence of legal NLP benchmarks for Cyrillic-script, civil-law jurisdictions. It was funded by SecondLayer.

\paragraph{Composition.} 2,000 Ukrainian court decisions (46~MB raw text) sampled from the 2024 partition of EDRSR, with 500 per jurisdiction (civil, criminal, commercial, administrative). Each decision includes full text, metadata labels, parsed facts section, extracted legal norms, and outcome labels. 110 result files contain 158K model predictions across 11 models.

\paragraph{Collection.} Decisions are sourced from EDRSR (\url{https://reyestr.court.gov.ua}), the official Ukrainian government registry of court decisions. All personal identifiers are anonymized at source (OSOBA\_1, ADRESA\_1). No human subjects were involved.

\paragraph{Preprocessing.} Facts sections extracted via rule-based parser matching canonical Ukrainian court decision structure (\foreignlanguage{ukrainian}{ВСТАНОВИВ\ldots ВИРІШИВ}). Legal norms extracted via regex, validated at 70\% agreement with Claude Sonnet 4.6. Cause categories mapped from 4,106 EDRSR codes to 22 macro-categories via keyword matching (58\% agreement with LLM judge).

\paragraph{Splits.} Fixed test set only (no train split). Few-shot examples drawn from a separate pool with deterministic seed (seed=42).

\paragraph{Distribution.} CC-BY-4.0 on HuggingFace (\url{https://huggingface.co/datasets/overthelex/ua-legal-bench}). Source decisions are public domain under Ukrainian law.

\paragraph{Maintenance.} Annual updates planned. Version-controlled with semantic versioning. Contact: \texttt{vladimir@legal.org.ua}.

\bibliographystyle{plainnat}

\end{document}